\documentclass[sigconf, 10pt]{acmart}
\renewcommand\footnotetextcopyrightpermission[1]{} 
\setcopyright{none}

\usepackage{algorithmic}
\usepackage{microtype}
\usepackage{graphicx}
\usepackage{booktabs} 
\usepackage{caption}
\usepackage{subcaption}
\usepackage{hyperref}       
\usepackage{url}            
\usepackage{amsmath}
\usepackage{multirow}
\usepackage{algorithm}

\usepackage{hyperref}

\newcommand{\Opt}{Acc-t-SNE~}

\newcommand{\beginsupplement}{%
        \setcounter{table}{0}
        \renewcommand{\thetable}{S\arabic{table}}%
        \setcounter{figure}{0}
        \renewcommand{\thefigure}{S\arabic{figure}}%
     }

\begin{document}

\title{Accelerating Barnes-Hut t-SNE Algorithm by Efficient Parallelization on Multi-Core CPUs}

\author[Chaudhary et. al.]{Narendra Chaudhary, Alexander Pivovar, Pavel Yakovlev, Andrey Gorshkov, Sanchit Misra}

\email{[narendra.chaudhary, alexander.pivovar, pavel.yakovlev, andrey.gorshkov, sanchit.misra]@intel.com}

\affiliation{%
  \institution{Intel Corporation}
  \streetaddress{}
  \city{}
  \state{}
  \country{}
  \postcode{}
}
\vskip 0.3in

\begin{abstract}

t-SNE remains one of the most popular embedding techniques for visualizing high-dimensional data. Most standard packages of t-SNE, such as scikit-learn, use the Barnes-Hut t-SNE (BH t-SNE) algorithm for large datasets. However, existing CPU implementations of this algorithm are inefficient. In this work, we accelerate the BH t-SNE on CPUs via cache optimizations, SIMD, parallelizing sequential steps, and improving parallelization of multithreaded steps.
Our implementation (\Opt) is up to $261\times$ and $4\times$ faster than scikit-learn and the state-of-the-art BH t-SNE implementation from daal4py, respectively, on a $32$-core Intel\textregistered Icelake cloud instance.

\textbf{Project Code URL:} \url{https://github.com/IntelLabs/Trans-Omics-Acceleration-Library/tree/master/applications/single_cell_pipeline}

\end{abstract}


\maketitle

\section{Introduction}
\label{Introduction}

Data analysis is the first step in many machine learning applications, and data visualization is a core component of data analysis. 
Visualization of complex high-dimensional data provides insights for hypothesis formation, algorithm choice, and hardware selection. Visualization techniques can also provide valuable insights into the inner workings of deep learning algorithms. Additionally, visualization techniques such as t-SNE \cite{van2008visualizing} and UMAP \cite{mcinnes2018umap} have found applications in large scale genomics data analysis. For example, t-SNE is used to visualize proximal cells in high-dimensional single-cell RNA-seq data \cite{10x2017transcriptional}.     

We can visualize a set of high-dimensional data points in two or three dimensions by using a dimensionality reduction technique. Dimensionality reduction techniques can be broadly classified into linear and non-linear techniques. Linear techniques, such as Principal Component Analysis (PCA), focus on keeping the low-dimensional representations of far-apart points.
Alternately, non-linear dimensionality reduction techniques, such as the t-SNE, focus on local distances and use neighborhood graphs to construct low-dimensional embeddings. 

Exact computation of t-SNE embeddings \cite{van2008visualizing} can be very slow. Therefore, Maaten et. al., \cite{van2014accelerating} proposed t-SNE acceleration using the tree-based Barnes-Hut\cite{barnes1986hierarchical} algorithm for computation of repulsive forces. Scikit-learn \cite{scikit-learn}, Multicore-TSNE \cite{Ulyanov2016}, and daal4py \cite{daal4py} provide parallel implementations of this algorithm for CPUs. Both Multicore-TSNE and daal4py t-SNE implementations are faster than scikit-learn for large datasets. To the best of our knowledge, daal4py t-SNE is the fastest implementation of BH t-SNE for CPUs. FIt-SNE \cite{linderman2019fast} uses an FFT-interpolation-based approach to compute further approximation of the repulsive force. FIt-SNE can be significantly faster, especially for large datasets. In addition to CPU implementations, there are several GPU accelerated t-SNE implementations \cite{burtscher2011efficient,chan2018t,chan2019gpu,nolet2022accelerating}. These approaches attempt to implement efficient GPU versions of BH t-SNE and FIt-SNE.

\subsection{Our Contributions} 
In this work, we start with the existing daal4py implementation of BH t-SNE consisting of the following major steps: (1) K Nearest Neighbors (KNN), (2) Binary Search Perplexity (BSP), (3) Quadtree building, (4) Summarization, (5) Attractive force computation, and (6) Repulsive force computation. KNN implementation in daal4py is already fairly efficient. We speedup the other steps on CPUs as follows.   
\begin{enumerate}
\item We adopt a Morton-code based algorithm to make quadtree building more efficient and parallelization friendly. Additionally, we develop a vectorized (SIMD) and multithreaded version of the quadtree building algorithm that scales significantly better than daal4py. 
\item Morton code based quadtrees, along with our efficient data layout, also improve single-thread performance of Summarization and Repulsive force computations.
\item We apply SIMD parallelism and software prefetching to the attractive force computation to improve the single-thread performance and multithread scaling. 
\item We accelerate the existing sequential implementations of BSP and Summarization by multithreading.
\end{enumerate}

Our single-thread improvements provide up to $2.6\times$ speedup over daal4py t-SNE. On a $32$-core Intel\textregistered IceLake based cloud instance, our implementation (\Opt) achieves up to $21\times$ speedup over single-threaded execution. This combination of improvements in single-thread performance and multi-thread scaling achieves up to $4\times$ speedup over state-of-the-art BH t-SNE implementation -- daal4py. Additionally, \Opt is up to $29\times$ faster than Multicore-TSNE, up to $3.8\times$ faster than FIt-SNE, and up to $261\times$ faster than scikit-learn t-SNE.
\section{Background}
\label{Background}

\subsection{Problem Statement}

Given the high dimensional input dataset $X = \{x_1, x_2, ..., x_N\}$ containing $N$ high-dimensional points $x_i$, we want to reduce it to two or three-dimensional embedding $Y = \{y_1, y_2, ..., y_N\}$ so that it can easily be visualized in a scatterplot. The goal is to preserve as much of the significant structure of the high-dimensional data as possible in the low-dimensional map. For brevity, we only discuss the case with a two-dimensional $Y$ in this paper as it is the most common use case.

\subsection{Brief overview of the Barnes Hut t-SNE Algorithm}

The t-distributed Stochastic Neighbor Embedding (t-SNE), first proposed by ~\cite{van2008visualizing, van2014accelerating}, is based on the Stochastic Neighbor Embedding (SNE) \cite{hinton2002stochastic} technique for data visualization and dimensionality reduction. It starts by converting the high-dimensional distances between all pairs of data points ($x_i$, $x_j$) into joint probabilities ($p_{ij}$) that represent similarities. Subsequently, it randomly initializes corresponding points in the low dimensional space and calculates joint probabilities ($q_{ij}$) between all pairs of data points ($y_i$, $y_j$). The algorithm aims to find the embeddings of the low dimensional points, such that the difference between $p_{ij}$ and $q_{ij}$ is minimized. The t-SNE algorithm consists of the following steps:

\subsubsection{Input similarity computation}
In the first step, the algorithm computes similarities ($p_{ij}$) between high-dimensional input point pairs ($x_i$, $x_j$) based upon distance between them ($d_{ij}$) using Gaussian kernels.

\begin{equation}
    p_{j|i} = \frac{exp(-d_{ij}^2 / 2\sigma_i^2)}{\sum_{k\neq i} exp(-d_{ik}^2 / 2\sigma_i^2)},  \qquad p_{i|i} = 0\\
\end{equation}
$$p_{ij} = \frac{p_{i|j} + p_{j|i}}{2N}$$

Here, $\sigma_i$ is the variance of the Gaussian kernel that is centered on data point $x_i$. These variances are set based upon a predefined input parameter called perplexity, denoted here by $u$. Values of $\sigma_i^2$ vary per object, and $\sigma_i^2$ tend to have smaller values in higher data density regions. The similarities ($p_{ij}$) form a joint probability distribution over all point pairs ($x_i, x_j$), and can be represented as an $N \times N$ similarity matrix $P$.

Similarities ($p_{ij}$) for distant points $i$ and $j$ can be extremely small. Therefore, \cite{van2014accelerating} proposed a sparse approximation of the similarity matrix ($P$). Sparse approximation accelerates the t-SNE algorithm without impacting the quality of the low dimension embedding. The algorithm obtains the sparse approximation by first finding $\lfloor 3u \rfloor$ nearest neighbors of each input point and then redefining the input similarities as following:

\begin{equation}
    p_{j|i} = \begin{cases}
    \frac{exp(-d_{ij}^2 / 2\sigma_i^2)}{\sum_{k \in N_i} exp(-d_{ik}^2 / 2\sigma_i^2)}, & \text{if $j \in N_i$} \\
    0, & \text{otherwise}
    \end{cases}
    \label{eq:perplexity}
\end{equation}
$$p_{ij} = \frac{p_{i|j} + p_{j|i}}{2N}$$

Here, $N_i$ are $\lfloor 3u \rfloor$ neighborhood points of $x_i$. Values of $\sigma_i^2$ are set such that the perplexity of the conditional distribution is equal to input perplexity $u$ via a binary search over $\sigma_i^2$. The nearest neighbors are found via the \textbf{k-nearest neighbors (KNN)} step and $\sigma_i^2$ by \textbf{binary search perplexity (BSP)} step.

Similarities $q_{ij}$ also get computed for low-dimensional embedding points $Y$. Here, the t-SNE algorithm uses the normalized Student-t kernel with a single degree of freedom to compute embedding space similarities $q_{ij}$.  

\begin{equation}
    q_{ij} = \frac{(1 + \|y_i - y_j\|^2)^{-1}}{\sum_{k\neq l}(1 + \|y_k - y_l\|^2)^{-1}}, \qquad q_{ii} = 0
\end{equation}

Similarities in embedding space $q_{ij}$ also form a joint-probability distribution over all embedding point pairs ($y_i, y_j$), and can be represented as an $N \times N$ similarity matrix $Q$.

\begin{figure*}[!tbh]
\vskip -0.1in
    \centering
    \begin{subfigure}{0.4\textwidth}
        \centering
        \includegraphics[width=\textwidth]{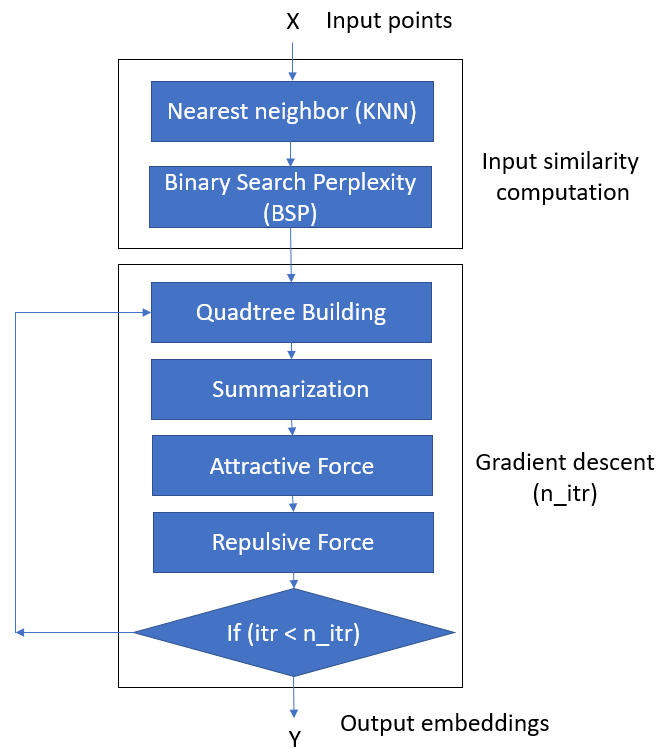}
        \caption{Major Steps of the BH t-SNE algorithm.}
        \label{fig:bhtsne_algo}
    \end{subfigure}%
    \hfill
    \begin{subfigure}{0.4\textwidth}
        \centering
        \includegraphics[width=\textwidth]{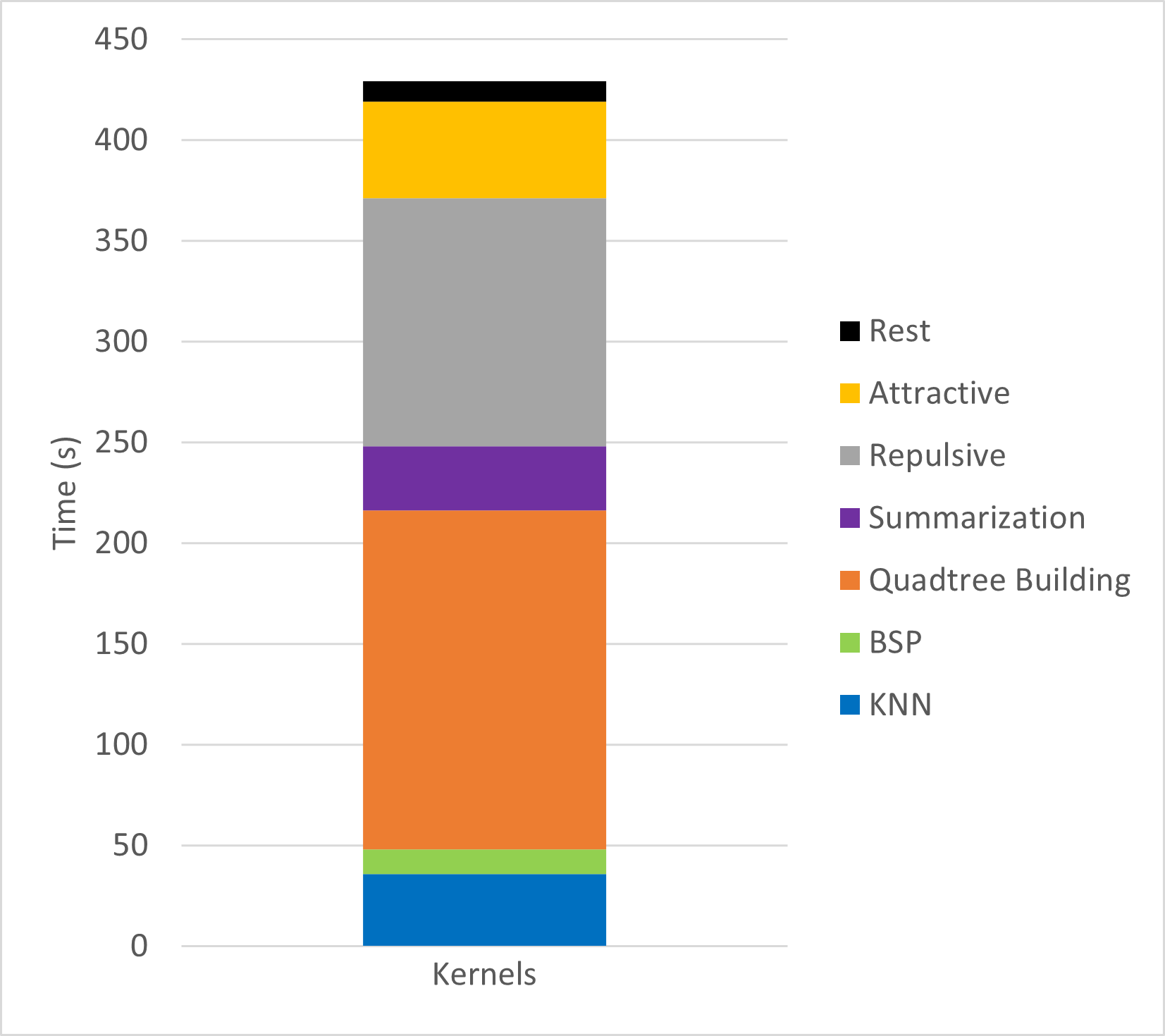}
        \caption{Profile of the BH t-SNE algorithm based implementation in daal4py v.2021.6 with a dataset of 1 million cells subsampled from the mouse brain cell dataset described in Section ~\ref{sec:datasets}}
        \label{fig:bhtsne_algo_profile}
    \end{subfigure}%
    \caption{The BH t-SNE algorithm and its profile.}
\end{figure*}

\subsubsection{Gradient Descent}
As a metric of difference between all the $p_{ij}$ and $q_{ij}$ values, t-SNE uses \textbf{Kulback-Leibler divergence (KL divergence)} since it helps to preserve the local structure of points. Therefore, minimizing the difference between $p_{ij}$ and $q_{ij}$ is the same as minimizing the KL divergence between the probability distributions of high-dimensional input space ($P$) and low-dimensional embedding space ($Q$). The algorithm uses gradient descent to minimize the KL divergence cost $C$.

\begin{equation}
    C(y) = KL(P\|Q) = \sum_{i \neq j} p_{ij} \log{\frac{p_{ij}}{q_{ij}}}
\end{equation}

By taking the gradient of the cost with respect to $y_i$, 
\begin{equation}
    \frac{\partial C}{\partial y_i} = 4 \sum_{i \neq j} (p_{ij} - q_{ij})q_{ij} Z(y_i - y_j)
\end{equation}

Where, $Z = \sum_{k\neq l}(1 + \|y_k - y_l\|^2)^{-1}$ is a normalization term. The cost gradient can be split into two parts,

\begin{equation}
    \frac{\partial C}{\partial y_i} = 4\left(\sum_{i \neq j} p_{ij} q_{ij} Z(y_i - y_j) - \sum_{i \neq j} q_{ij}^2 Z(y_i - y_j)\right)
\end{equation}

Here, the first summation can be interpreted as the attractive force $F_{attr}$ from all points on $y_i$, and the second summation can be interpreted as the repulsive force $F_{rep}$ from all points.
\begin{equation}
\frac{\partial C}{\partial y_i} = 4(F_{attr} + F_{rep})
\end{equation}

\paragraph{Attractive Force Computation:}
By recognising, $q_{ij} Z(y_i - y_j) = (1 + \|y_i - y_j\|^2)^{-1}$,
\begin{equation}
F_{attr} = \sum_{i \neq j} \frac{p_{ij}}{(1 + \|y_i - y_j\|^2)}
\label{eq:attractive}
\end{equation}.
Therefore, attractive force computation a $y_i$ point can be done by performing the above summation over all non-zero elements from $i_{th}$ row of the sparse matrix $P$. 

\paragraph{Repulsive Force Computation:}
The exact computation of all repulsive forces $F_{rep}$ has $O(N^2)$ computational complexity. Therefore, doing it at every gradient descent step is expensive. 

Maaten et. al. ~\cite{van2014accelerating} proposed accelerating the t-SNE repulsive force computation to $O(N \ log(N))$ time complexity by doing an approximate computation of $F_{rep}$ using the \textbf{Barnes-Hut algorithm} \cite{barnes1986hierarchical}. 
The Barnes-Hut method leverages the insight that, for any point, points that are distant in low dimensional space and are close to each other, exert nearly the same force. This method has three steps, 1) \textbf{Quadtree building} from current $Y$, 2) \textbf{Summarization} and 3) \textbf{Repulsive Force} computation via Quadtree traversal in $O(N \ log(N))$ time. We provide more details of these steps in Section ~\ref{sec:methods_quadtree}.

In summary, Figure~\ref{fig:bhtsne_algo} illustrates the major steps of the BH t-SNE algorithm -- 1) K-nearest neighbors (KNN), 2) Binary Search Perplexity (BSP), and for each iteration of gradient descent -- 3) Quadtree building, 4) Summarization, 5) Attractive force computation, and 6) Repulsive force computation.

\subsection{Profile of BH t-SNE algorithm}

We use daal4py \cite{daal4py} implementation of BH t-SNE for profiling since, based on our experiments, that is the fastest implementation prior to our work. We obtain the profile on a $32$-core CPU described in Table~\ref{tab:sys-config} using one of the largest datasets described in Section~\ref{sec:datasets}. Figure~\ref{fig:bhtsne_algo_profile} shows that BH t-SNE has a relatively flat profile, thus requiring acceleration of all the steps in order to achieve end-to-end acceleration.

\section{Our Acceleration of BH t-SNE}
\label{methods}


\begin{table}[t]
\caption{Summary of improvements in each step.}
\label{tab:optimizations}
\vspace{0.1in}
\begin{center}
\begin{small}
\begin{sc}
\begin{tabular}{cc}
\toprule
Step & Improvements \\
\midrule
BSP & multithreading with Numba \\
\hline
Quadtree  & Morton codes to make\\
Building  & parallelization friendly, \\
  & multithreading, \\
  & data layout improvements \\
\hline
Summarization & multithreading \\
\hline
Repulsive &  data layout improvement \\
\hline
Attractive & SIMD, software prefetching \\
\bottomrule
\end{tabular}
\end{sc}
\end{small}
\end{center}
\end{table}

We use the implementation of BH t-SNE in daal4py \cite{daal4py} v.2021.6.0 as our baseline implementation and build our accelerated implementation on top of that. 
Given the flat profile, we accelerate most of the major steps of the algorithm (shown in Figure ~\ref{fig:bhtsne_algo}) with additional emphasis on steps that are more time consuming. For example, since BSP takes a very small percentage of the time, we accelerate it by multi-threading the existing Python source code in daal4py using Numba~\cite{10.1145/2833157.2833162}. On the other hand, steps such as quadtree building, attractive force computation, repulsive force computation, consume large percentages of time. For these steps, we worked on the existing C++ implementation under daal4py and used rigorous methods such as 1) modifying the algorithm to make it more parallelization friendly, 2) data layout changes to improve data locality, 3) multi-threading, 4) software prefetching to alleviate memory latency bottleneck and 4) SIMD to reduce number of instructions. We provide a summary of our improvements to each step in Table ~\ref{tab:optimizations} and more details in the following Subsections.    

\subsection{K Nearest Neighbours} 
We use the existing KNN implementation from daal4py for our optimized code. Daal4py version of KNN is fairly efficient and scales well with the number of cores as shown in Section ~\ref{sec:kernel-scaling} and Figure ~\ref{fig:kernel-scaling}a. 

\subsection{Binary Search Perplexity} 
This step computes the conditional similarities $p_{j|i}$ for all input point pairs based upon equation~\ref{eq:perplexity}. For each $i$, it computes the equation~\ref{eq:perplexity} multiple times with different values of $\sigma_i^2$ until the perplexity of the conditional distribution $p_{j|i}$ is equal to input target perplexity $u$. This search for the value of $\sigma_i^2$ is performed through a binary search where in each step, $\sigma_i^2$ is the middle point of successively smaller intervals. Prior implementations of this BSP are single-threaded. We have parallelized this step by recognizing that we can compute $p_{j|i}$ for each $i$ independently in parallel. We use Numba's \cite{10.1145/2833157.2833162} parallel range utility to multi-thread this step. 

\subsection{Quadtree Building}
\label{sec:methods_quadtree}

\begin{figure}[!tbh]
\vskip -0.1in
  \centering
  \includegraphics[width=0.48\textwidth]{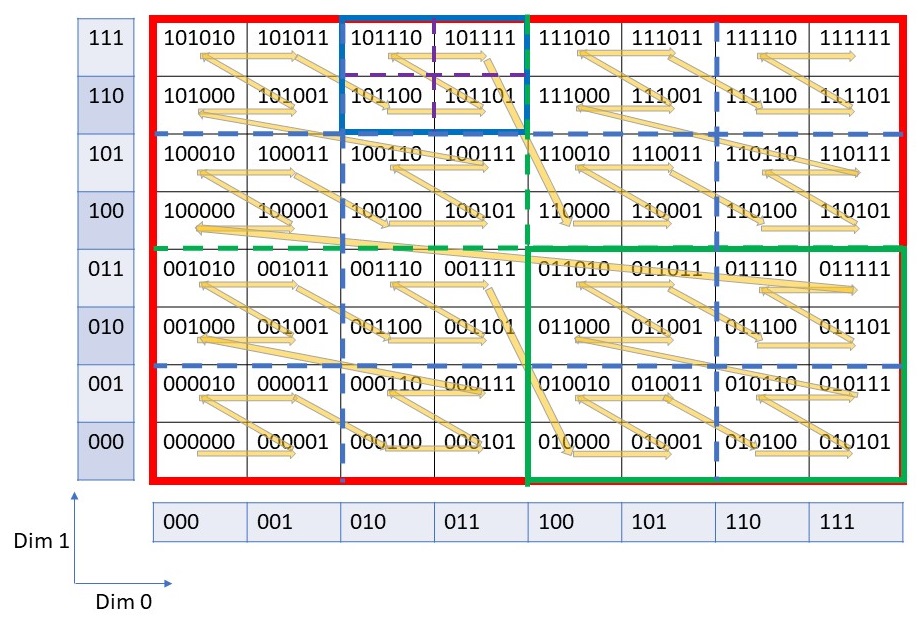}
  \vspace{-0.1cm}
  \caption{Two-dimensional space split into hierarchical cells. Binary numbers are Morton codes associated with cells. The increasing order of morton code follows the orange line (Z-order).
  }
\label{fig:morton}
\end{figure}

\begin{figure*}[!tbh]
\vskip -0.1in
  \centering
  \includegraphics[width=0.8\textwidth]{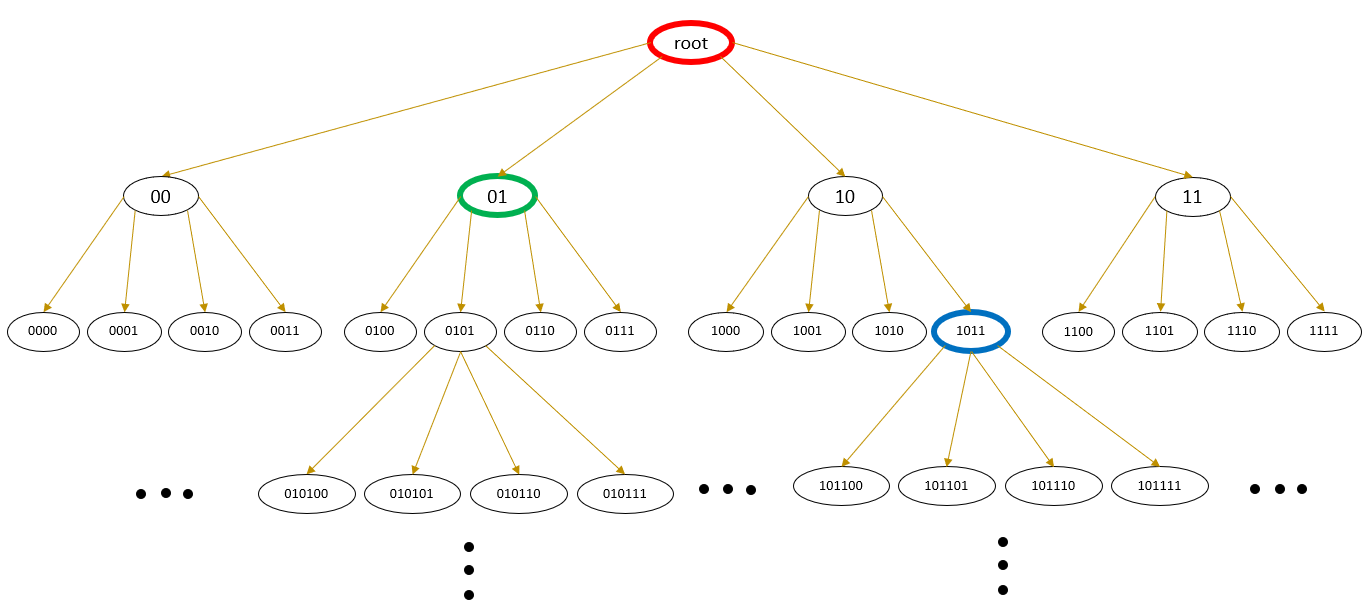}
  \caption{Quadtree representation of two-dimensional space. Each non-leaf node represents a cell. Binary numbers are Morton codes associated with cells.}
  \label{fig:quadtree}
\end{figure*}

At every gradient step of the BH t-SNE algorithm, we need to build a new quadtree to compute repulsive forces $F_{rep}$. 
The boundaries of the output two-dimensional space are defined by the minimum/maximum values in $Y$ along the two dimensions. The BH method partitions this two-dimensional space into four quadrants creating four rectangular cell. These cells are recursively partitioned into four sub quadrants each creating smaller cells until 1) there is only a single point in a cell or 1) the cells are too small. The root node of the quadtree represents the entire two-dimensional space. Each non-leaf node represents a rectangular cell of the two-dimensional space and has four children corresponding to its four subquadrants. Leaf nodes represent a cell containing a
single point. Figure~\ref{fig:morton} shows a two-dimensional space partitioned into cells, and Figure~\ref{fig:quadtree} shows the quadtree associated with those cells. For illustration, we color some cells in Figure~\ref{fig:morton} and corresponding nodes in Figure~\ref{fig:quadtree} to show the relationship. 

Daal4py constructs the quadtree by going through the various levels of the tree starting from the root level. At each level, it goes through each node of the tree and checks for the corresponding cell needs to be partitioned into quadrants. If a cell needs partitioning, it creates four children of the corresponding node and divides all the points in that cell into the four quadrants. This algorithm is computationally very expensive since each time a cell is partitioned, we need to go over all the points in the cell to split them across the quadrants. So, each point is traversed as many times as the depth of the tree for that point. Instead, similar to ~\cite{burtscher2011efficient}, we use Morton codes that allows us to traverse each point only once. Morton codes are also useful for making the algorithm more parallelization friendly.

\begin{algorithm}[tb]
  \caption{Morton Code Formation Algorithm}
  \label{alg:morton_algo}
\begin{algorithmic}[1]
  \STATE {\bfseries Input:} Embedding points $y_i = (y_i[0], y_i[1]), \forall \ i \in \{0, N-1\}$, $(cent[0], cent[1])$, $r_{span}$
  \STATE {\bfseries Output:} 64-bit Morton codes $M_i, \forall \ i \in \{0, N - 1\}$ 
  \STATE Initialize $M_i = 0, \forall \ i \in \{0, N - 1\}$.
  \STATE $(y_{root}[0], y_{root}[1]) \gets (cent[0] - r_{span}, cent[1] - r_{span})$
  \STATE $scale = 2^{31} / r_{span} $
  \
  \FOR{$i=0$ {\bfseries to} $N-1$ in parallel}
      \STATE $(m[0], m[1]) \gets (y_i[0] - y_{root}[0], y_i[1] - y_{root}[1]) * scale$
      \newline
      \FOR{$d=0$ {\bfseries to} $1$} 
      \STATE $m[d] \gets m[d] \ \AND \ (\mathtt{0x00000000ffffffff})$
      \STATE $m[d] \gets m[d] \ \OR (m[d] \ll 16) $
      \STATE $m[d] \gets m[d] \ \AND (\mathtt{0x0000ffff0000ffff})$
      \STATE $m[d] \gets m[d] \ \OR (m[d] \ll 8) $
      \STATE $m[d] \gets m[d] \ \AND (\mathtt{0x00ff00ff00ff00ff})$
      \STATE $m[d] \gets m[d] \ \OR (m[d] \ll 4) $
      \STATE $m[d] \gets m[d] \ \AND (\mathtt{0x0f0f0f0f0f0f0f0f})$
      \STATE $m[d] \gets m[d] \ \OR (m[d] \ll 2) $
      \STATE $m[d] \gets m[d] \ \AND (\mathtt{0x3333333333333333})$
      \STATE $m[d] \gets m[d] \ \OR (m[d] \ll 1) $
      \STATE $m[d] \gets m[d] \ \AND (\mathtt{0x5555555555555555})$
      \newline
      \ENDFOR
      \newline
      \STATE $M_i \gets m[0] \ \OR (m[1] \ll 1)$
  \ENDFOR
\end{algorithmic}
\end{algorithm}

\textbf{Morton Codes: } Morton codes \cite{morton1966computer} are spatial location codes \cite{van1996spatial} that map multi-dimensional data into one-dimension while preserving the locality of data points. We form Morton codes by bitwise interleaving of the coordinates in the two dimensions. For example, dimension $0$ value of $3 = 011_b$ and dimension $1$ value of $7 = 111_b$ has Morton code of $47 = 101111_b$ (see Figure~\ref{fig:morton}).

Morton codes or Z-order codes have several properties due to which they are used in quadtree construction \cite{van1996spatial}. In general, they are used in construction of bounded volume hierarchies \cite{lauterbach2009fast}. Sorted Morton codes ensure that the data points that are clustered close to each other in two-dimension are also close to each other in the sorted order. Therefore, they get stored close together in a one-dimensional data structure. Additionally, a quadtree cell/node can be represented as a linear range of Morton codes. For example, in Figure~\ref{fig:morton} the blue colored cell can be represented as a range $ \{101100_b, ..., 101111_b \}$ and the green colored cell can be represented as a range $ \{010000_b, ..., 011111_b \}$. The longest common prefix of the Morton codes in a cell is enough to identify tree level and cell (see Figure~\ref{fig:quadtree}). Depth-first traversal can also be understood as selecting subranges recursively from the full range of Morton codes.
We use 64-bit Morton codes that can represent $2^{64}$ cells. First, we convert each two-dimensional point into a Morton code by using Algorithm~\ref{alg:morton_algo}. The algorithm computes Morton codes from maximum span radius ($r_{span}$) of 2D space and its center point $(cent[0], cent[1])$. We can construct each embedding point's Morton code independently. Therefore, codes or all $N$ embedding points can be constructed in parallel. We use both SIMD parallelism -- automatically done by the compiler due to the simple structure of the algorithm --  and explicit multi-threading. 

\textbf{Parallel Quadtree Building: } At any level of the tree, all the nodes and the subtrees rooted at those nodes can be processed in parallel. One way to parallelize quadtree building is to distribute the nodes at each level across threads. However, many levels near the top and bottom of the tree may not have sufficient nodes to keep all the threads busy. Moreover, nodes created by a thread may be processed for further partitioning in the next level by another thread hampering data locality and causing data to jump from cache of one core to another. Therefore, we use a different approach. First, we create the tree to the minimum level such that there are sufficiently large number of nodes at that level. Subsequently, we distribute the nodes at that level across threads. Each thread is responsible not only for processing the assigned nodes but also creating the entire subtrees rooted at those nodes. Since different subtrees may be of different sizes -- thus requiring different amount of work, we use dynamic thread scheduling over the nodes. For dynamic thread scheduling, the number of nodes need to be sufficiently larger than the number of threads. We also store all the nodes of a tree at a particular level in a contiguous manner to aid data locality.

\subsection{Parallel Summarization} 

After the quadtree is built, the summarization step computes the center-of-mass for each cell in the constructed quadtree for future use in the repulsion step. The current daal4py implementation of the summarization step is single-threaded, and therefore it can take a significant amount of time for large-scale datasets (see Figure~\ref{fig:bhtsne_algo_profile}). In the summarization step, we traverse the quadtree from the bottom up to compute the center-of-mass of all cells. For leaf nodes, the mass is always one.
For non-leaf nodes, the center-of-mass gets computed based on the number of points and their positions inside the cell. To compute the center-of-mass of a higher-level node, we only need the center-of-mass of its four children and the count of points in them. Therefore, we can compute the center-of-mass of all the cells at one level of the quadtree in parallel. 
Hence, we start from the bottom level, processing all the nodes in parallel, and then proceed to the level above it.      

\subsection{Repulsive Force Computation}
While computing $F_{rep}$ for a point, if a cell is sufficiently far and sufficiently small, BH method approximates the contribution of all the points in the cell with the summary of the cell.
To compute repulsive force on a point, the BH method traverses the quadtree in a depth-first search (DFS) manner. At each node of the tree, it checks if the summary at the node can be used as approximation instead of going further down the subtree. The summary is used if the following condition, proposed by \cite{barnes1986hierarchical}, is true:
\begin{equation}
    \frac{r_{cell}}{\|y_i - y_{cell}\|^2} < \theta
\end{equation}
Where $\|y_i - y_{cell}\|^2$ is the square of the distance to the cell, $r_{cell}$ is the radius of the cell and $\theta$ is an input parameter that trades off speed and accuracy.

The DFS of the tree can lead to irregular memory access leading to memory latency bottlenecks if the nodes of the tree are randomly located. However, due to the structured data locality of our quadtree, the DFS experiences significantly better data locality, thereby accelerating the repulsive step.

\subsection{Attractive Force Computation}

\begin{algorithm}[tb]
  \caption{Attractive force algorithm in daal4py}
  \label{alg:attractive_daal}
\begin{algorithmic}[1]
  \STATE {\bfseries Input:} Embedding points $y_i = (y_i[0], y_i[1]), \forall \ i \in \{0, N-1\}$, Sparse similarity CSR matrix $P = (row, col, val)$
  \STATE {\bfseries Output:} $F_{attr} = (attr_i[0], attr_i[1]), \forall \ i \in \{0, N - 1\}$ 
  \FOR{$i=0$ {\bfseries to} $(N-1)$ in parallel}
    \STATE $(attr_i[0], attr_i[1]) \gets (0, 0)$
    \FOR{$ind=row[i]$ {\bfseries to} $(row[i+1] - 1)$ }
        \STATE $j \gets col[ind] - 1$
        \STATE $PQ \gets \frac{val[ind]}{(1 + (y_i[0] - y_j[0])^2 + (y_i[1] - y_j[1])^2)}$
        \STATE $attr_i[0] \gets attr_i[0] + PQ * (y_i[0] - y_j[0]) $
        \STATE $attr_i[1] \gets attr_i[1] + PQ * (y_i[1] - y_j[1])$
    \ENDFOR
  \ENDFOR
\end{algorithmic}
\end{algorithm}

Equation~\ref{eq:attractive} computes the attractive force on a embedding point $y_i$ by all other points $y_j$. 
The input similarities $p_{ij}$ are computed only once, before the start of gradient descent steps. In approximate similarity computation \cite{van2014accelerating} the similarities $p_{ij}$ are only calculated for set of $\lfloor 3u \rfloor$ nearest neighbors $N_i$ of each input N. Therefore, the similarity matrix, $P$ is a sparse matrix with few non-zero values and is stored in compressed sparse row (CSR) format. The calculation of $F_{attr}$ for each $y_i$ -- equivalent to processing the $i$-th row of the CSR matrix -- is independent and can be done in parallel. The daal4py implementation of Algorithm~\ref{alg:attractive_daal} takes this approach and scales well to multiple cores, as shown in Figure~\ref{fig:kernel-scaling-daal4py}. Therefore, we focus on improving the single-threaded performance of the attractive step. 
We can observe from the innermost for loop of Algorithm~\ref{alg:attractive_daal} that, for each $y_i$, all its neighbors $y_j$ are read. There are $\lfloor 3u \rfloor$ such $y_j$ values spread over an array of $N$ points in a pseudo random fashion. Given $\lfloor 3u \rfloor << N$, accessing $y_j$ could lead to irregular memory accesses.
This can cause misses in caches of different levels depending upon the size of $N$. Larger values of $N$ will cause more cache misses causing CPU stalls. We alleviate this by software prefetching the $y_j$ values of a later $y_i$ while we are processing the current $y_i$. Better cache locality also result in better multithread scaling.

In addition to software prefetching, we also improve the single-threaded performance via SIMD parallelization of the innermost loop. Since the datatype is double-precision floating point, we hand-vectorize the innermost loop with AVX512 to process eight loop iterations simultaneously. We use AVX512 instructions for load, store and several compute operations. We also need to use AVX512 gather operation for gathering eight $y_i$ values. There is sufficient compute in the innermost loop so that, despite the high latency of gather operations, we see healthy performance improvement due to vectorization.

\section{Experiments and Results}
\label{results}

In this section, we first detail our experimental setup and datasets used. Subsequently, we present results on comparison of end-to-end execution time and accuracy of our implementation with other implementations. Our results show that we achieve significant performance gain without compromising accuracy. Further, we deep dive into the various dimensions of performance improvement - single thread performance and multicore scaling - for end-to-end execution and individual steps. 

\subsection{Experimental setup}

\begin{table}[t]
\caption{System Configuration.}
\label{tab:sys-config}
\vspace{0.1in}
\begin{center}
\begin{small}
\begin{sc}
\begin{tabular}{lr}
\toprule
 & AWS EC2 \\
 & \MakeLowercase{c6i.16xlarge}\\
\midrule
Cores / CPU & $32$\\
Threads / Core & $2$\\
AVX register width (bits) & $512$, $256$, $128$\\
Vector Processing Units (VPU) & $2$/Core\\
Base Clock Frequency (GHz) & $2.9$ \\
L1D, L2 Cache (MiB) & $1.5$, $40$\\
L3 Cache (MiB) / CPU & $54$\\
DRAM (GiB) / CPU & $256$\\
\bottomrule
\end{tabular}
\end{sc}
\end{small}
\end{center}
\vskip -0.2in
\end{table}

All of the experiments presented in this paper are conducted on AWS EC2 c6i.16xlarge cloud instances as briefed in Table~\ref{tab:sys-config}. The instance provides a $32$ core 3rd generation Intel\textregistered \hspace{0.01cm} Xeon\textregistered \hspace{0.01cm} Platinum 8375C @$2.90$ GHz base clock frequency and $256$ {\tt GB} memory over a single NUMA domain. For more details, visit ~\cite{aws-c6i}.
All the experiments use a single thread per core. Following the evaluation methodology used in \cite{van2014accelerating}, we run each implementation for $1000$ iterations. Additionally, we keep input parameter values same as the default parameters of scikit-learn t-SNE. 

\subsection{Datasets}
\label{sec:datasets}
We conduct experiments with six datasets of different types, sizes, and input dimensions. These standard datasets are readily obtainable from internet for experimentation. \\ 
\textbf{Mouse brain cell dataset:} The mouse brain cell dataset \cite{10x2017transcriptional} contains single-cell RNA sequencing (scRNA-seq) data for 1.3 million mouse brain cells. The original raw dataset contains 27998 features for each cell. However, in a single-cell pipeline, t-SNE usually gets computed after data preprocessing and principal component analysis (PCA) steps. We run the t-SNE algorithm after PCA, with 20 principal components as features. The final dataset contains 1291337 data points, each with 20 features or dimensions. \\
\textbf{Digit dataset: } The Digit dataset from scikit-learn \cite{scikit-learn} contains 1797 grayscale images of size $8 \times 8 = 64$ pixels. Each image contains a handwritten digit between $\{0, 1, ...,9\}$. \\
\textbf{MNIST dataset:} The MNIST handwritten digits dataset \cite{deng2012mnist} contains 70000 grayscale images of size $28 \times 28 = 784$ pixels. \\ 
\textbf{CIFAR\-10 dataset:} The CIFAR-10 dataset \cite{krizhevsky2009learning} contains 60000 color images of size $32 \times 32$ pixels. The dataset consists of images of ten different classes or objects. The dataset has 3072 ($32 \times 32 \times 3$) features or input dimensions. \\ 
\textbf{Fashion MNIST dataset:} The fashion MNIST dataset \cite{xiao2017fashion} contains 70000 grayscale images of size $28 \times 28 = 784$ pixels. The dataset contains images of fashion products from 10 different categories. \\
\textbf{SVHN dataset: } The street view house numbers (SVHN) dataset \cite{netzer2011reading}, we use contains 99289 color images of size $32 \times 32$ pixels. The dataset contains google street view images of house numbers and has ten different classes. \\

Except the mouse brain cell dataset, we apply t-SNE algorithm directly on input points and collect execution time and divergence results. Supplementary Figures \ref{fig:vis_logits}, \ref{fig:vis_mnist}, \ref{fig:vis_fashion_mnist}, \ref{fig:vis_mouse}, \ref{fig:vis_cifar_10}, and \ref{fig:vis_svhn} show the visualization plots for these datasets with different t-SNE implementations.

\subsection{Comparison of end-to-end performance with state-of-the-art}

\begin{figure*}[!tbh]
\vskip -0.05in
    \centering
    \begin{subfigure}{0.33\textwidth}
        \centering
        \includegraphics[width=\textwidth]{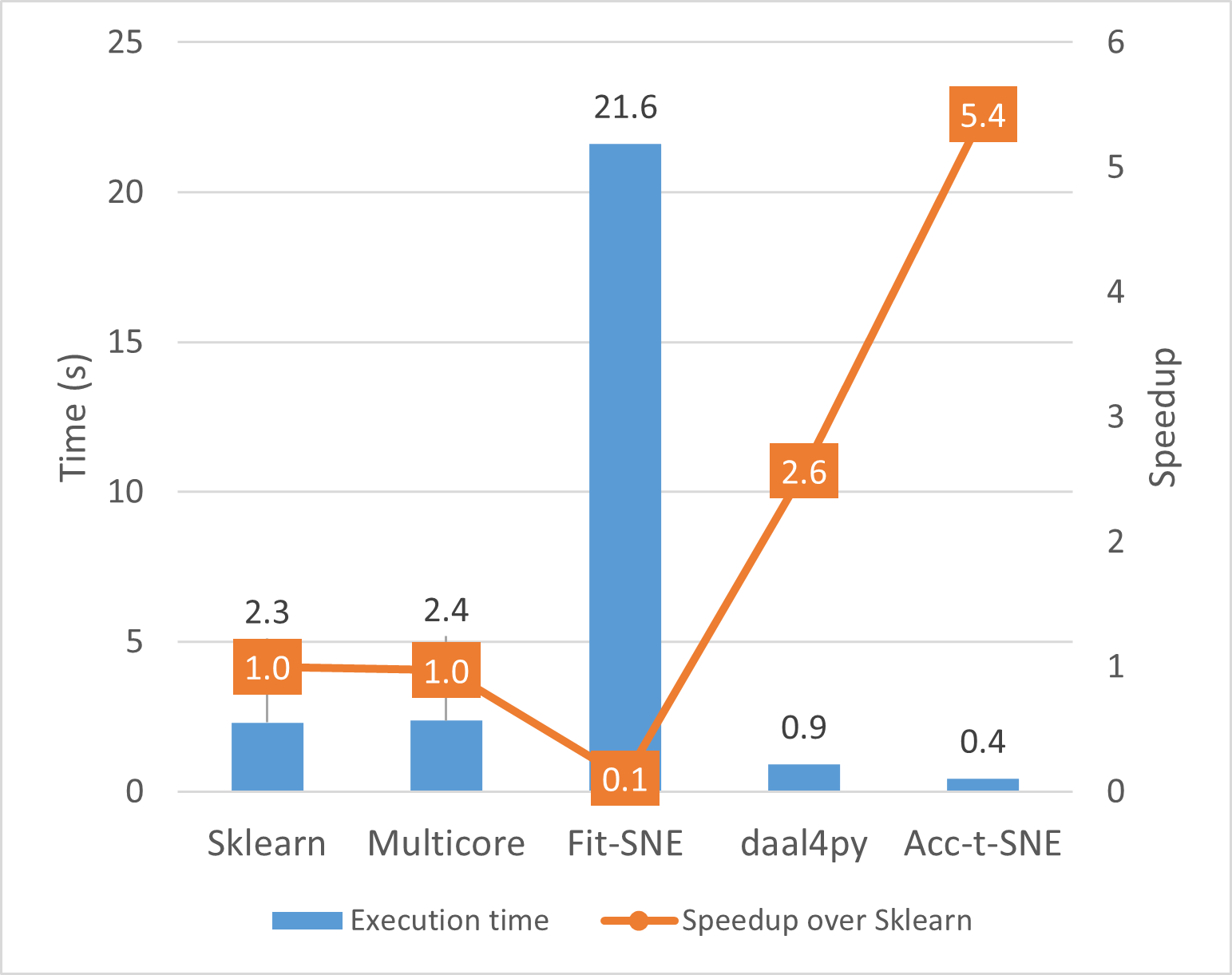}
        \caption{Digit dataset.}
    \end{subfigure}%
    \hfill
    \begin{subfigure}{0.33\textwidth}
        \centering
        \includegraphics[width=\textwidth]{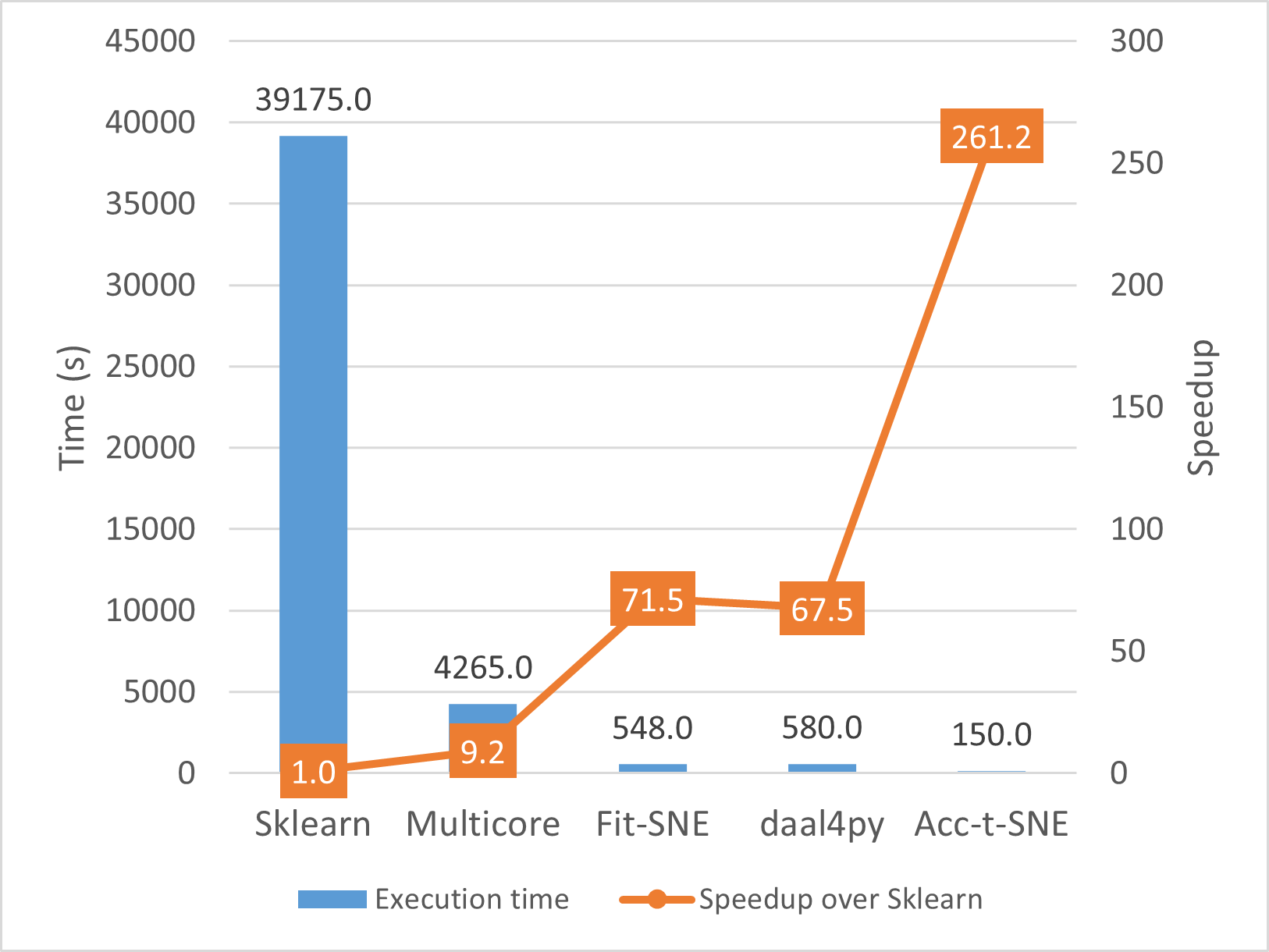}
        \caption{1.3 million mouse brain cell.}
    \end{subfigure}%
    \hfill
    \begin{subfigure}{0.33\textwidth}
        \centering
        \includegraphics[width=\textwidth]{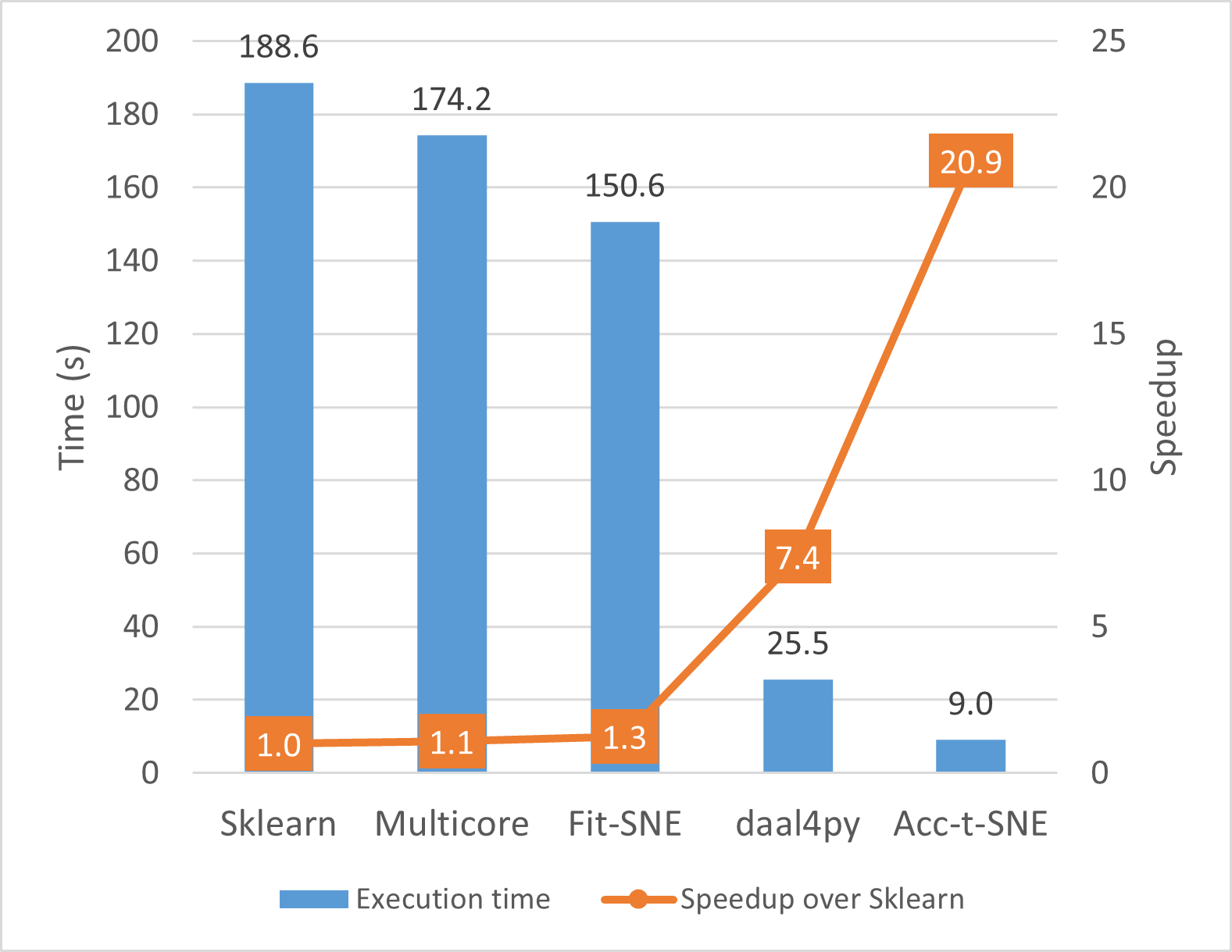}
        \caption{MNIST dataset.}
    \end{subfigure}%
    \hfill
    \begin{subfigure}{0.33\textwidth}
        \centering
        \includegraphics[width=\textwidth]{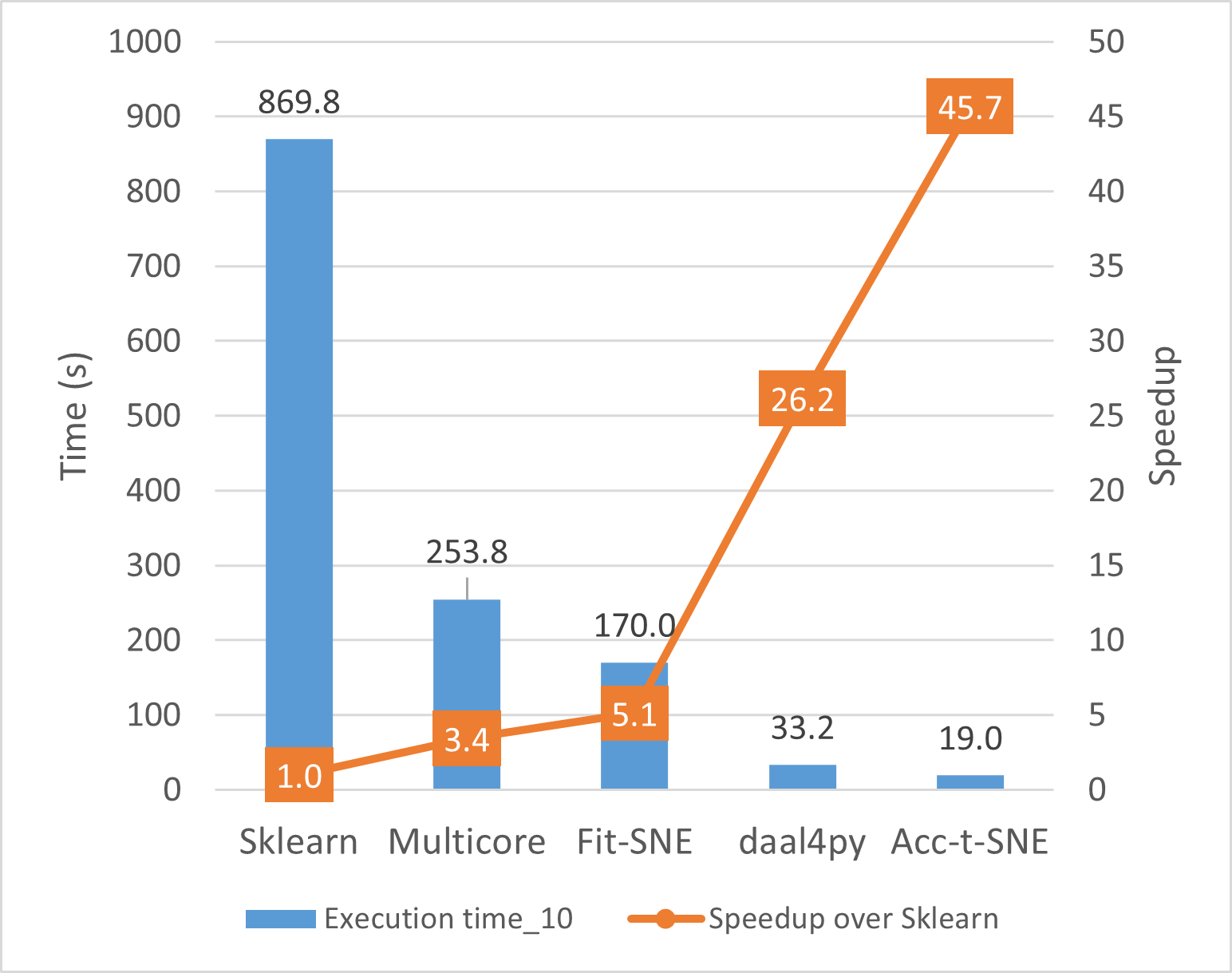}
        \caption{CIFAR\_10 dataset.}
    \end{subfigure}%
    \hfill
    \begin{subfigure}{0.33\textwidth}
        \centering
        \includegraphics[width=\textwidth]{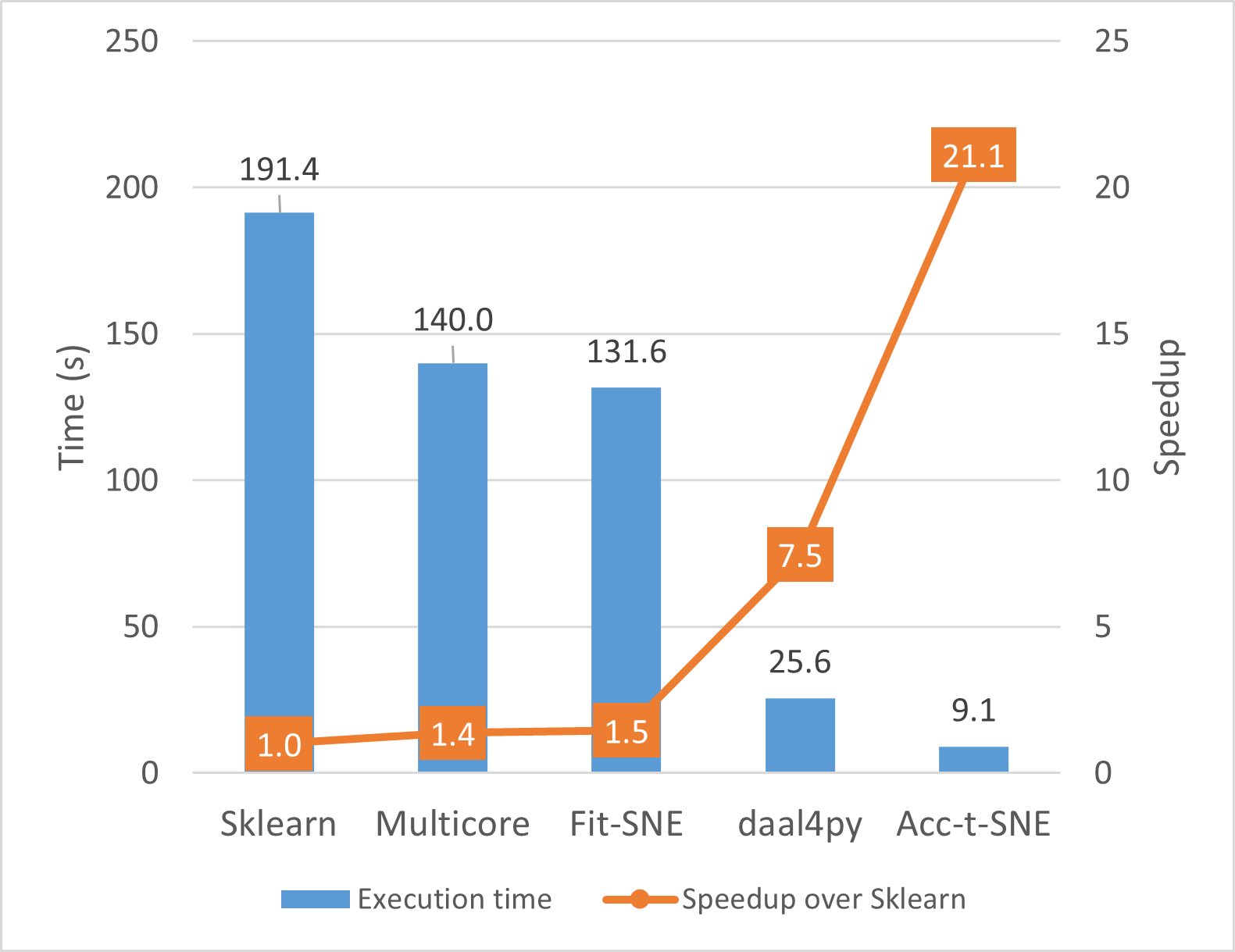}
        \caption{Fashion MNIST dataset.}
    \end{subfigure}%
    \hfill
    \begin{subfigure}{0.33\textwidth}
        \centering
        \includegraphics[width=\textwidth]{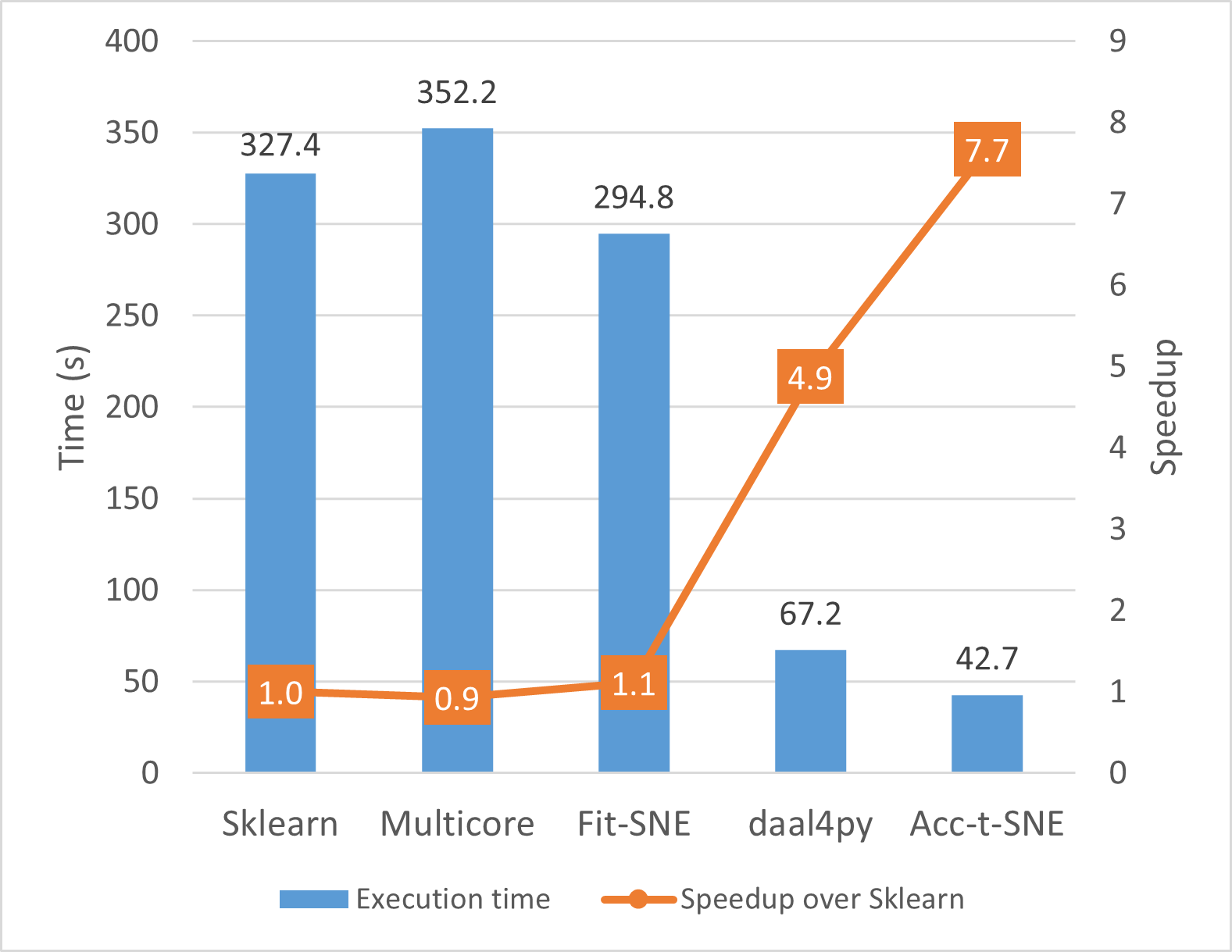}
        \caption{SVHN dataset.}
    \end{subfigure}%
    \caption{End-to-end performance comparison of the various implementations using all $32$ cores over six standard datasets. The bars represent the execution time for each implementation, while the line represents the speedup of each implementation over Sklearn.}
    \label{fig:e2e-perf-comp}
\end{figure*}

In this section, we compare the end-to-end execution time of \Opt for several standard datasets with the state of the art implementations: scikit-learn t-SNE (Sklearn) ~\cite{scikit-learn}, multicore t-SNE (Multicore) ~\cite{Ulyanov2016}, FFT-accelerated Interpolation-based t-SNE (FIt-SNE) ~\cite{linderman2019fast} and the t-SNE implementation in daal4py v2021.6.0 \cite{daal4py}. We run all of these algorithm in double-precision (float64) data format for consistency and fair comparison. 

Figure~\ref{fig:e2e-perf-comp} presents the comparison. Since the implementation in scikit-learn is the most widely used, we present the speedup of various implementations over that. \Opt is consistently the fastest, achieving speedups of $5.4\times - 261.2\times$ over scikit-learn. Speedup on a particular dataset depends on a variety of factors such as dataset size, input dimension size, etc. In general, larger datasets have more opportunity of parallelism due to wider quadtrees, and therefore, the speedup is higher for larger datasets. Additionally, we can also run the \Opt implementation in single-precision (float32) data format and achieve upto $1.6\times$ speedup over the double-precision implementation without significant loss of accuracy (see Supplementary Table~\ref{tab:float32}).  

\subsection{Evaluation of accuracy}

\begin{table}[t]
\caption{Comparison of accuracy for all the six datasets. The lowest KL-divergence values for each dataset are highlighted in bold font.}
\label{tab:divergence}
\vspace{0.1in}
\begin{center}
\begin{small}
\begin{sc}
\begin{tabular}{cccc}
\toprule
Dataset & Sklearn & daal4py & Optimized \\
\midrule
Digit    & \textbf{0.740} & 0.853 & 0.853 \\
Mouse-1.3M & 10.237 & \textbf{7.064} & 7.280 \\
MNIST    & 3.233 & \textbf{3.175} & 3.196 \\
CIFAR\_10 & 4.369 & \textbf{4.357} & 4.374 \\
Fashion MNIST & 2.989 & \textbf{2.947} & 2.967 \\
SVHN      & 4.305 & \textbf{4.283} & 4.387 \\
\bottomrule
\end{tabular}
\end{sc}
\end{small}
\end{center}
\vskip -0.2in
\end{table}

In this section, we evaluate the accuracy of \Opt. We compared our accuracy against two implementations: 1) scikit-learn, since it is the most widely used, and 2) daal4py, since Figure~\ref{fig:e2e-perf-comp} shows that it is the fastest implementation prior to our work. We use the standard metric of accuracy comparison - KL-divergence that measures the difference between probability distributions of the input space and the embedding space. As can be seen from Table~\ref{tab:divergence}, the KL-divergence values of \Opt are very close to the smallest KL-divergence value for each dataset apart from the Digit dataset.

\subsection{Evaluation of end-to-end performance improvements}

For these experiments, we pick the largest dataset - 1.3 million mouse brain cell dataset. We first evaluate the single-thread end-to-end performance of t-SNE implementations. After that, we evaluate the scaling performance of these implementation by increasing number of cores.

\subsubsection{End-to-end single thread performance}

\begin{table}[t]
\caption{Performance comparison of single threaded runs of various implementations. Dataset used: 1.3 million mouse brain cell dataset. First column and second column mention the implementation name and the execution time in seconds, respectively. The last column presents speedup of an implementation over Sklearn.}
\label{tab:implementation-wise-single-thread}
\vskip 0.1in
\begin{center}
\begin{small}
\begin{sc}
\begin{tabular}{lrr}
\toprule
Implementation & Execution & Speedup\\
 & Time (s) & \\
\midrule
Sklearn & $28818$ & $1.0\times$ \\
Multicore & $15973$ & $1.8\times$ \\
FIt-SNE & $3077$ & $9.4\times$ \\
daal4py & $7684$ & $3.8\times$ \\
\Opt & $3125$ & $9.2\times$ \\
\bottomrule
\end{tabular}
\end{sc}
\end{small}
\end{center}
\vskip -0.1in
\end{table}

Table~\ref{tab:implementation-wise-single-thread} shows the single-thread performance of various t-SNE implementations on 1.3 million mouse cell dataset. FIt-SNE has the fastest single-threaded implementation, with \Opt being a close second. However, \Opt is $2.5\times$ faster than the second fastest BH t-SNE implementation from daal4py. 

\subsubsection{End-to-end multicore scaling}
\label{sec:end-to-end-scaling}

\begin{figure}[!tbh]
\centering
\includegraphics[width=0.4\textwidth]{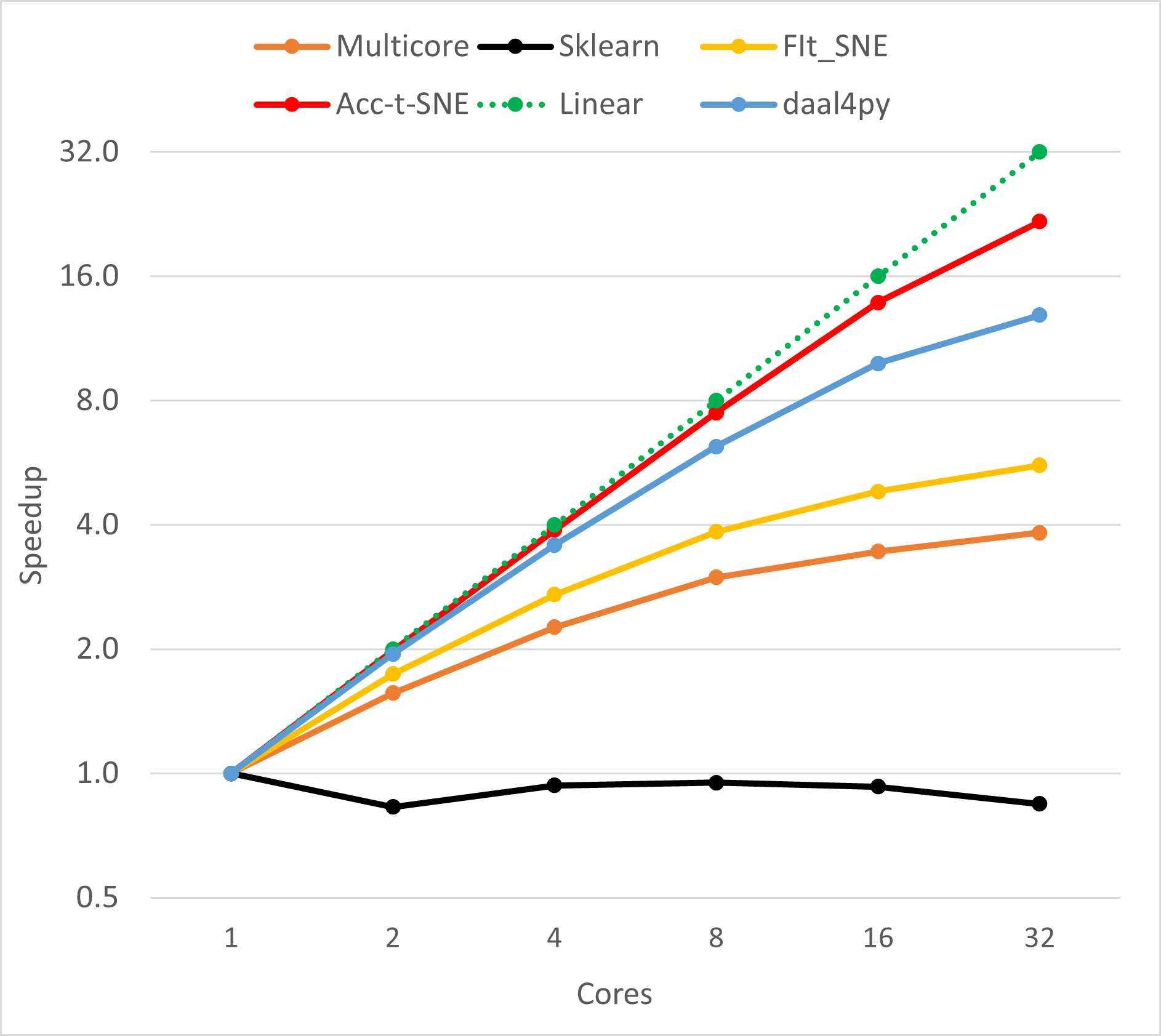}
    \caption{Scaling of various implementations with respect to the number of cores on the mouse brain cell data. The dotted line represents linear scaling. Each solid line presents the scaling of one of the implementations. The points in each solid line show the speedup of the corresponding implementation with respect to its own single-thread execution.}
    \label{fig:multicore-scaling}
\end{figure}

Figure~\ref{fig:multicore-scaling} presents the end-to-end multicore scaling of various implementations of t-SNE for the 1.3 million mouse brain cell dataset. \Opt scales better than all the other implementations. It achieves the best speedup of $22\times$ when using $32$ cores with respect to the single-core performance. Clearly, the improvements in \Opt help it scale better than other implementations. This is the scaling of end-to-end execution time. Different steps may scale differently and may have seen different improvements in multicore scaling. In the next section, we will dive deeper into how the improvements help scale individual steps better. As we observed in the previous section, FIt-SNE does well on a single thread but scales poorly to multiple cores. If it could scale well then it can close the gap with \Opt.

\subsection{Evaluating performance improvements of individual steps}
\label{sec:kernel-scaling}
This section presents the performance improvements of \Opt for individual steps. For these experiments, for the sake of brevity, we present the results using a single dataset: 1 million cells sampled from mouse-brain cell dataset. We pick daal4py for comparison since it is the fastest implementation prior to our work. First, we present the performance improvements of \Opt over daal4py on a single thread. Subsequently, we present our improved scaling.
As mentioned in Section ~\ref{methods}, the KNN implementation in daal4py is efficient and scales well. We have just used the KNN from daal4py in \Opt.

\subsubsection{Single thread performance}

\begin{table}[t]
\caption{Performance comparison on a single thread of daal4py and \Opt for each step. Dataset used: 1 million cells subsampled from mouse cell dataset. The first column mentions the step name, second and third columns present the runtime in seconds for each step. The last column presents the speedup of \Opt over daal4py for each step.}
\label{tab:kernel-wise-single-thread}
\vskip 0.1in
\begin{center}
\begin{small}
\begin{sc}
\begin{tabular}{lrrr}
\toprule
Step & daal4py & \Opt & Speedup\\
 & Time (s) & Time (s) & \\
\midrule
BSP & $12.4$ & $12.2$ & $1.0\times$ \\
Tree Building & $174.4$ & $39.0$ & $4.5\times$ \\
Summarization & $29.3$ & $5.6$ & $5.3\times$ \\
Attractive & $1226.0$ & $568.5$ & $2.2\times$ \\
Repulsive & $3016.3$ & $501.6$ & $6.0\times$ \\
\midrule
Total time & $5391.8$ & $2048.1$ & $2.6\times$ \\
\bottomrule
\end{tabular}
\end{sc}
\end{small}
\end{center}
\vskip -0.1in
\end{table}

Table~\ref{tab:kernel-wise-single-thread} demonstrates that we achieve $2.2-6\times$ speedup over daal4py for individual steps, leading to $2.6\times$ reduction in total time. This performance improvement is a result of 1) efficient Morton-code based algorithm in Quadtree building, 2) vectorization and software prefetching in the Attractive force step, 3) improvement in data locality and tree traversal by changing the data layout of the tree in Summarization and Repulsive force step. We have not made any single thread performance improvement in BSP since the percentage of time spent in that step is already very small.

\subsubsection{Multicore scaling}
\label{sec:multicore-scaling}

\begin{figure}[!tbh]
    \centering
    \begin{subfigure}{0.4\textwidth}
        \centering
        \includegraphics[width=\textwidth]{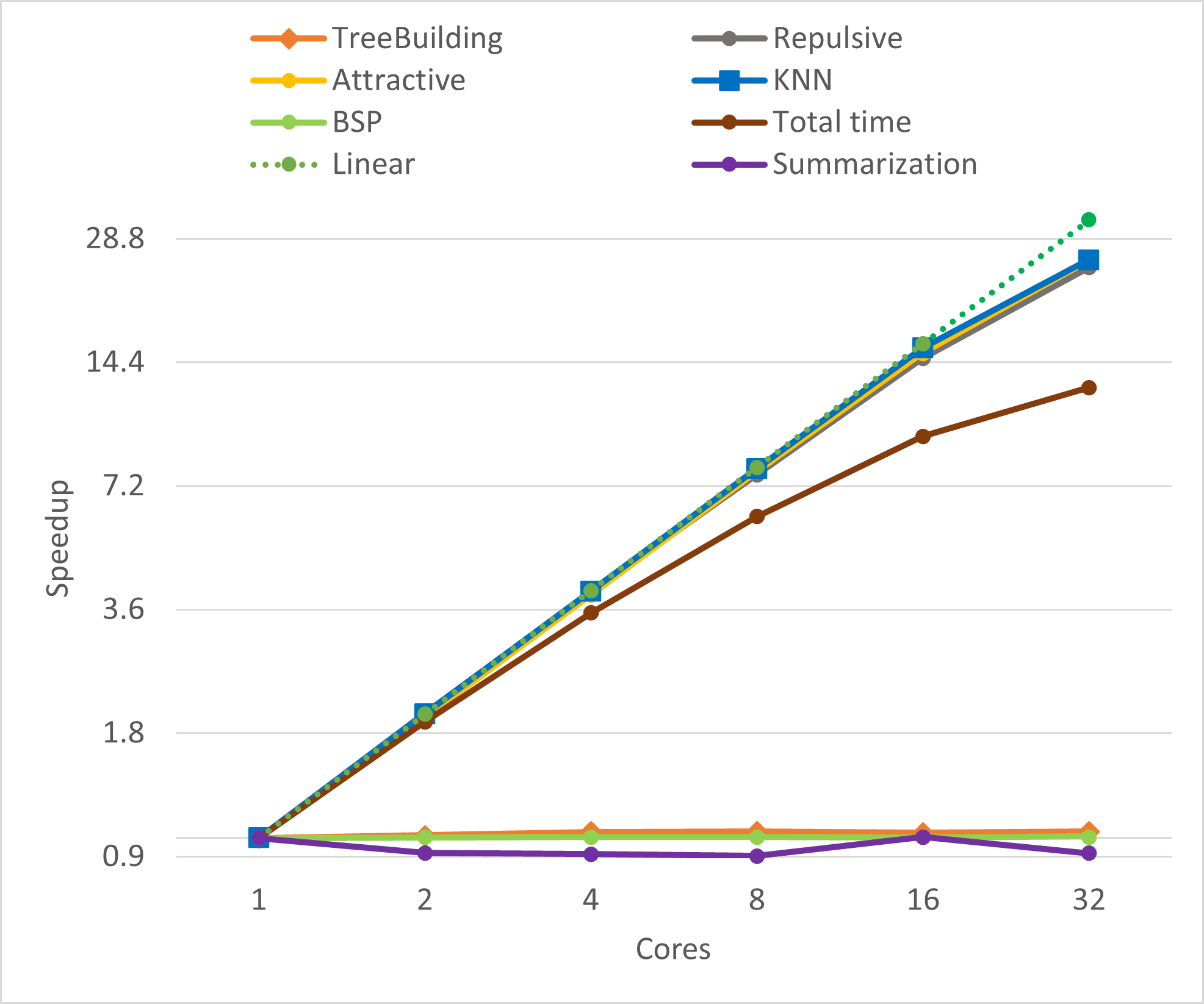}
        \caption{Scaling of steps in daal4py.}
        \label{fig:kernel-scaling-daal4py}
    \end{subfigure}%
    \hfill
    \begin{subfigure}{0.4\textwidth}
        \centering
        \includegraphics[width=\textwidth]{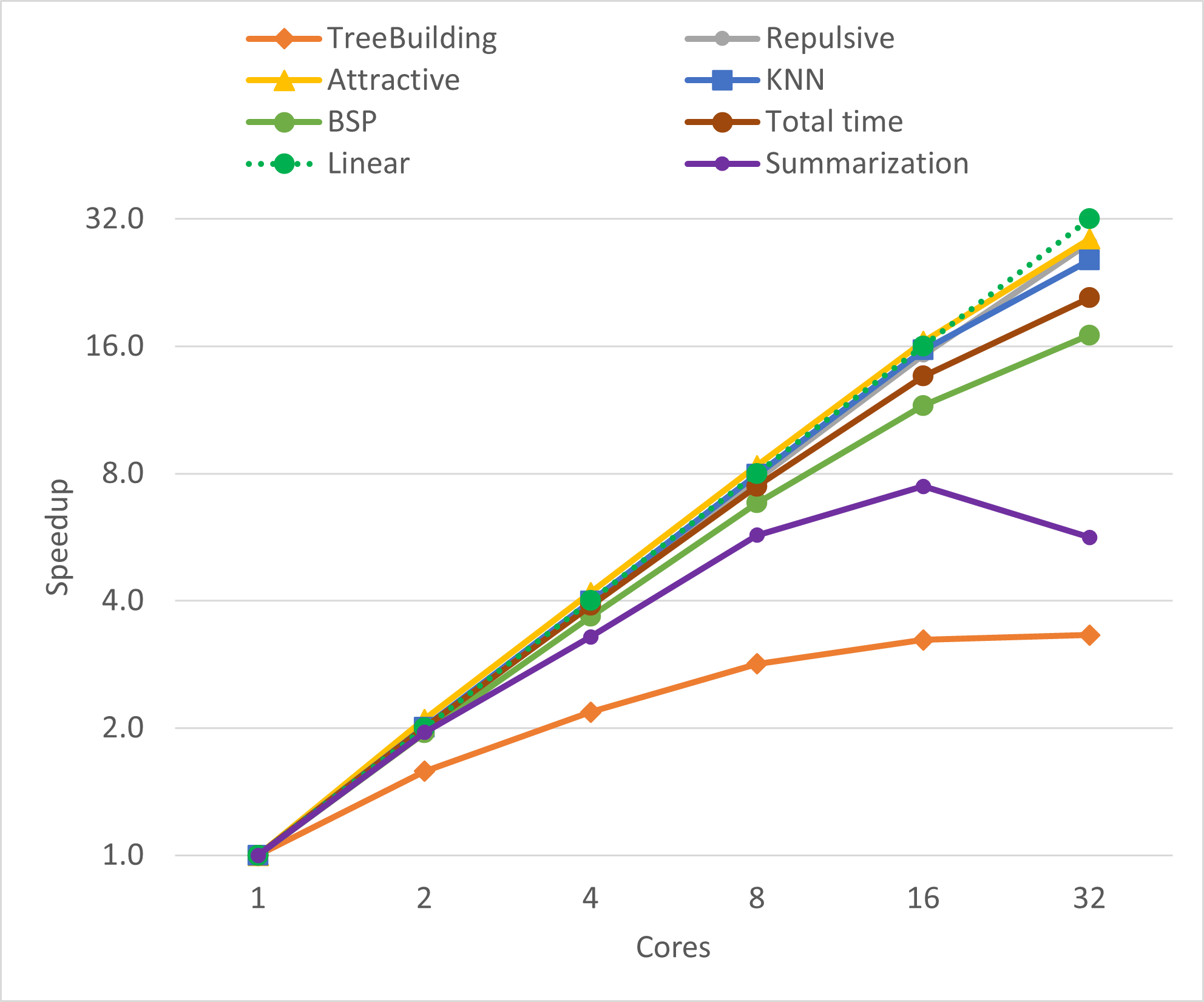}
        \caption{Scaling of steps in \Opt.}
        \label{fig:kernel-scaling-opt}
    \end{subfigure}%

    \caption{Scaling of daal4py and \Opt for each step. Dataset used: 1 million cells subsampled from mouse-brain cell dataset. The dotted line represents linear scaling. Each solid line presents the scaling of one of the steps. The points in each solid line present the speedup of the corresponding step with respect to its own single thread execution.}
    \label{fig:kernel-scaling}
\end{figure}

The results on multicore scaling of individual steps are presented in Figure~\ref{fig:kernel-scaling}. \Opt shows significant improvements in multicore scaling of various steps. These improvements provide additional performance gains over the speedups obtained on a single core. Note the following about the speedup of $32$ cores over a single core for each step for both implementations:

\begin{itemize}
    \item The speedup of Attractive step increases from $24\times$ for daal4py to $28.7\times$ for \Opt.
    \item The speedup of Repulsive step increases from $26.8\times$ for daal4py to $28.1\times$ for \Opt.
    \item The speedup of Summarization step increases from $1.1\times$ for daal4py to $5.7\times$ for \Opt. 
    \item Daal4py gets no speedup for Summarization, Quadtree building, and BSP since these steps are single threaded. On the other hand, \Opt achieves $5.7\times$ speedup for Summarization, $3.3\times$ for Quadtree building and $17\times$ speedup for BSP. 
\end{itemize}

\subsubsection{Multicore performance}

\begin{table}[t]
\caption{Performance comparison on entire 32 cores of daal4py and \Opt for each step. Dataset used: 1 million cells subsampled from mouse cell dataset. The first column mentions the step name, second a third columns present the runtime in seconds for each step. The last column presents the speedup of \Opt over daal4py for each step.}
\label{tab:kernel-wise-multi-thread}
\vskip 0.1in
\begin{center}
\begin{small}
\begin{sc}
\begin{tabular}{lrrr}
\toprule
Step & daal4py & \Opt & Speedup\\
 & Time (s) & Time (s) & \\
\midrule
BSP & $12.3$ & $0.7$ & $17\times$\\
Tree Building & $168.3$ & $11.7$ & $14.3\times$\\
Summarization & $31.9$ & $1.0$ & $32.4\times$\\
Attractive & $48.0$ & $19.8$ & $2.4\times$\\
Repulsive & $123.0$ & $17.8$ & $6.9\times$\\
\midrule
Total time & $431.1$ & $98.3$ & $4.4\times$\\
\bottomrule
\end{tabular}
\end{sc}
\end{small}
\end{center}
\vskip -0.1in
\end{table}

Table~\ref{tab:kernel-wise-multi-thread} demonstrates the combined effect of single thread improvements and multicore scaling improvements done in \Opt. We achieve $2.4-32.4\times$ speedup over daal4py for individual steps, leading to $4.4\times$ reduction in the total time. Considering that Attractive and Repulsive were the primary steps in the single threaded execution of daal4py, the performance improvements in those steps have certainly helped. However, the $14.3\times$ speedup achieved for Quadtree building is quite significant since, with Attractive and Repulsive steps achieving good multicore scaling, the Quadtree step becomes the most time consuming step of daal4py in the $32$-core execution.
The performance improvements of BSP and Summarization steps are also significant. Otherwise, with the performance improvements in other steps, they would have become bottlenecks.

\section{Conclusion}
\label{Conclusion}

Efficient implementations of machine learning algorithms enable researchers to conduct large-scale experiments. In this work, we have implemented an accelerated version (\Opt) of the t-SNE algorithm for modern multi-core CPUs. We employed several improvements of single-thread performance and multithread scaling to a majority of steps of BH t-SNE and implemented parallel versions of the previously sequential steps. \Opt provides superior single-thread performance and multi-core scaling compared to the state-of-the-art BH t-SNE implementations. To the best of our knowledge, it is the fastest implementation of t-SNE on CPUs providing upto $4\times$ speedup over previous state-of-the-art.

\bibliographystyle{ACM-Reference-Format}
\bibliography{tSNE}

\noindent{\small Optimization Notice: Software and workloads used in performance tests may have been optimized for performance only on Intel microprocessors. Performance tests, such as SYSmark and MobileMark, are measured using specific computer systems, components, software, operations and functions. Any change to any of those factors may cause the results to vary. You should consult other information and performance tests to assist you in fully evaluating your contemplated purchases, including the performance of that product when combined with other products. For more information go to http://www.intel.com/performance. Intel, Xeon, and Intel Xeon Phi are trademarks of Intel Corporation in the U.S. and/or other countries.}
\appendix

\beginsupplement
    
\section{Supplemental material}

\begin{figure*}[!tbh]
  \centering
  \includegraphics[width=0.99\textwidth]{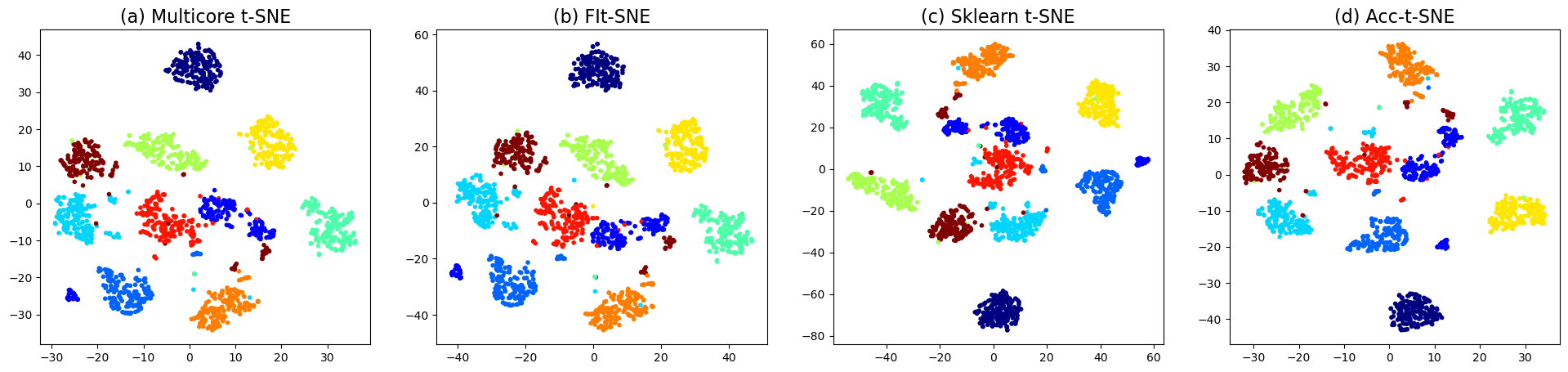}
  \caption{Digits t-SNE plots. }
  \label{fig:vis_logits}
\end{figure*}

\begin{figure*}[!tbh]
  \centering
  \includegraphics[width=0.99\textwidth]{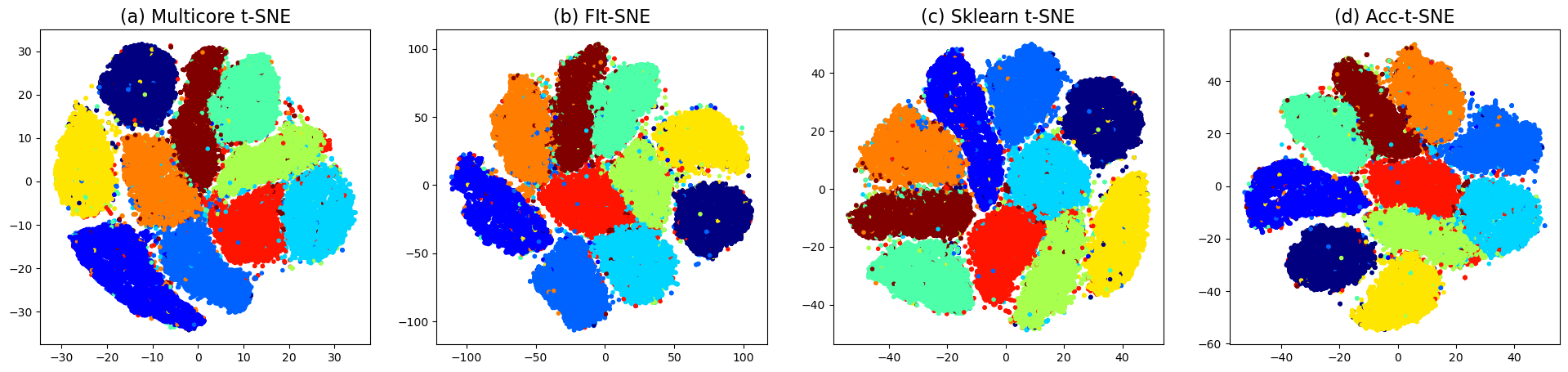}
  \caption{MNIST t-SNE plots. }
  \label{fig:vis_mnist}
\end{figure*}

\begin{figure*}[!tbh]
  \centering
  \includegraphics[width=0.99\textwidth]{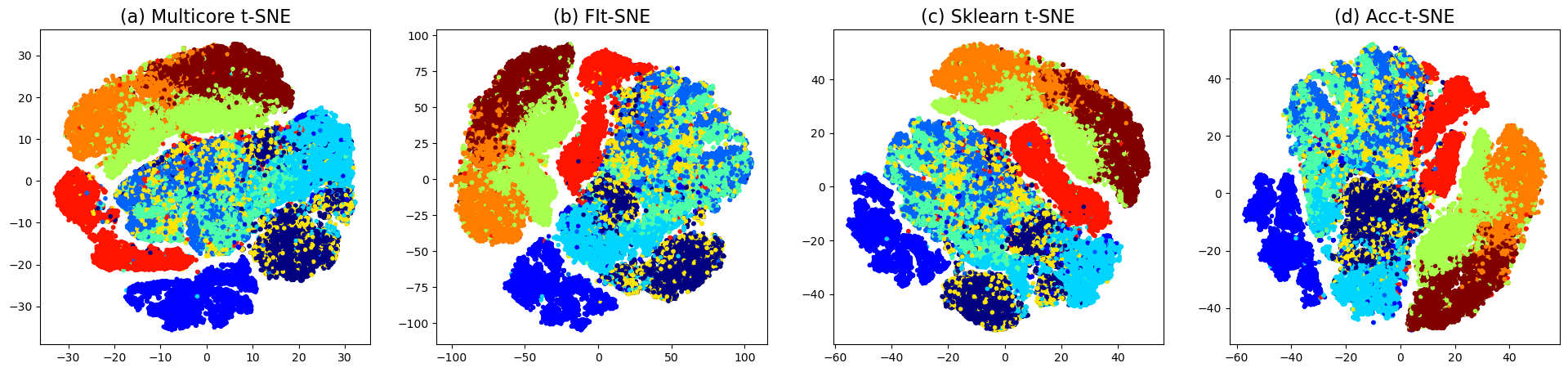}
  \caption{Fashion MNIST t-SNE plots. }
  \label{fig:vis_fashion_mnist}
\end{figure*}

\begin{figure*}[!tbh]
  \centering
  \includegraphics[width=0.99\textwidth]{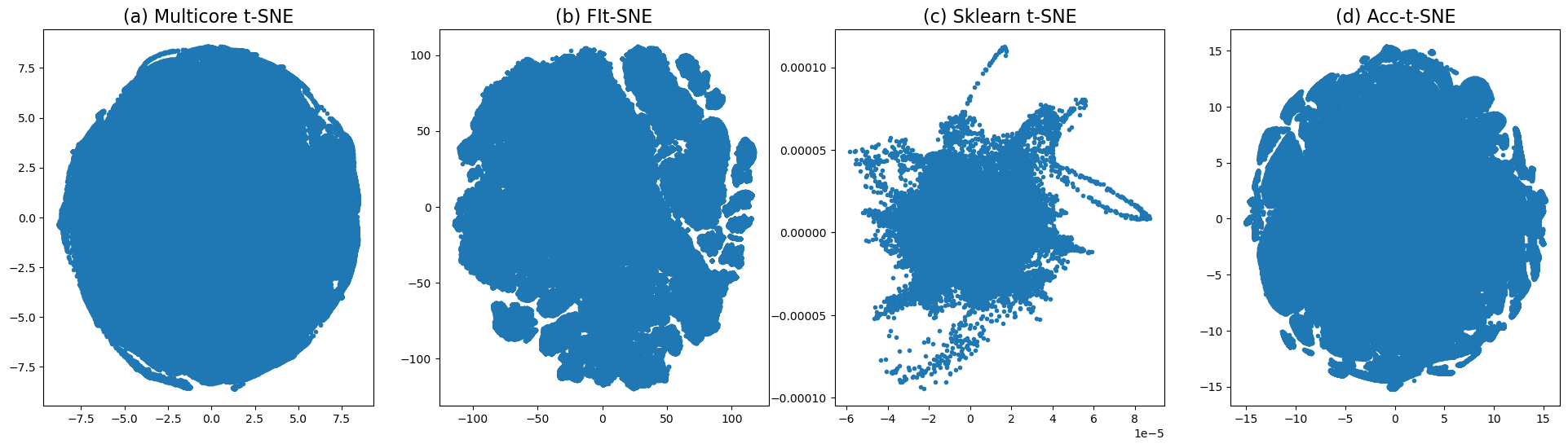}
  \caption{1.3 million mouse-brain cell t-SNE plots. }
  \label{fig:vis_mouse}
\end{figure*}

\begin{figure*}[!tbh]
  \centering
  \includegraphics[width=0.99\textwidth]{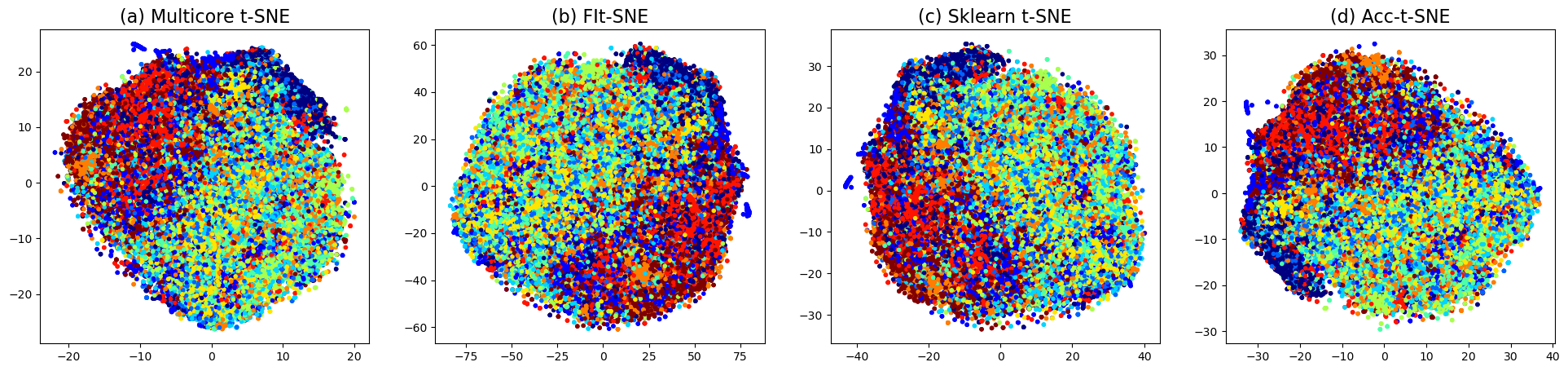}
  \caption{CIFAR-10 t-SNE plots. }
  \label{fig:vis_cifar_10}
\end{figure*}

\begin{figure*}[!tbh]
  \centering
  \includegraphics[width=0.99\textwidth]{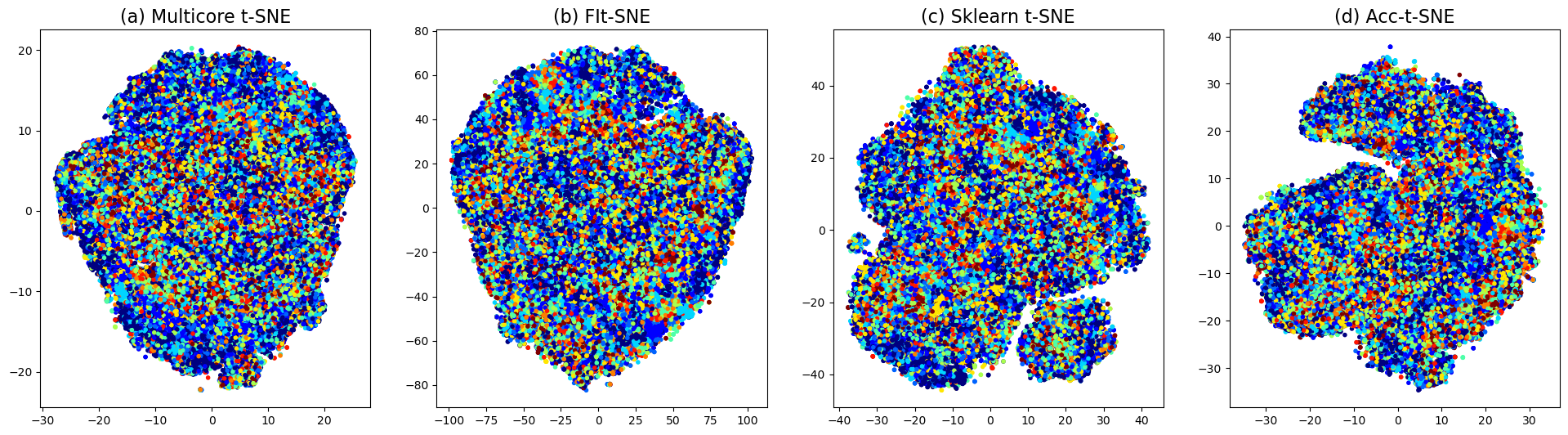}
  \caption{SVHN t-SNE plots. }
  \label{fig:vis_svhn}
\end{figure*}

\begin{table*}[t]
\caption{\Opt computation in single-precision (Float32) data format. }
\label{tab:float32}
\vskip 0.1in
\begin{center}
\begin{small}
\begin{sc}
\begin{tabular}{lccccr}
\toprule
Dataset & Time (s) & Divergence & Time (s) & Divergence & Speedup \\
 & (float32) & (float32) & (float64) & (float64) &  \\
\midrule
Digit  & 0.4 & 0.857 & 0.4 & 0.853 & $0.99\times$\\
Mouse-1.3M & 100 & 7.261 & 144 & 7.280 & $1.4\times$ \\
MNIST    & 6.4 & 3.196 & 9.0 & 3.196 & $1.4\times$\\
CIFAR\_10 & 11.7 & 4.375 & 19.0 & 4.374 & $1.6\times$\\
Fashion MNIST & 6.3 & 2.967 & 9.1 & 2.967 & $1.4\times$\\
SVHN      & 26.0 & 4.386 & 42.7 & 4.387 & $1.6\times$\\
\bottomrule
\end{tabular}
\end{sc}
\end{small}
\end{center}
\vskip -0.1in
\end{table*}

\end{document}